\documentclass{article}

\usepackage{PRIMEarxiv}

\usepackage[utf8]{inputenc} 
\usepackage[T1]{fontenc}    
\usepackage{hyperref}       
\usepackage{url}            
\usepackage{booktabs}       
\usepackage{amsfonts}       
\usepackage{nicefrac}       
\usepackage{microtype}      
\usepackage{lipsum}
\usepackage{fancyhdr}       
\usepackage{graphicx}       
\graphicspath{{media/}}     
\usepackage{bm}
\usepackage{amsmath}
\usepackage{float}
\usepackage{bbding}
\usepackage{makecell}
\usepackage[perpage]{footmisc}

\title{IP-Adapter: Text Compatible Image Prompt Adapter for Text-to-Image Diffusion Models
}

\author{
  Hu Ye,  Jun Zhang{\hypersetup{hidelinks}\thanks{Corresponding author}} , Sibo Liu,  Xiao Han,  Wei Yang \\
  Tencent AI Lab \\
  \texttt{\{huye, junejzhang, siboliu, haroldhan, willyang\}@tencent.com} \\
}

\begin{document}
\maketitle

\begin{abstract}
Recent years have witnessed the strong power of large text-to-image diffusion models for the impressive generative capability to create high-fidelity images. However, it is very tricky to generate desired images using only text prompt as it often involves complex prompt engineering. An alternative to text prompt is image prompt, as the saying goes: "an image is worth a thousand words". Although existing methods of direct fine-tuning from pretrained models are effective, they require large computing resources and are not compatible with other base models, text prompt, and structural controls. In this paper, we present IP-Adapter, an effective and lightweight adapter to achieve image prompt capability for the pretrained text-to-image diffusion models. The key design of our IP-Adapter is decoupled cross-attention mechanism that separates cross-attention layers for text features and image features. Despite the simplicity of our method, an IP-Adapter with only 22M parameters can achieve comparable or even better performance to a fully fine-tuned image prompt model. As we freeze the pretrained diffusion model, the proposed IP-Adapter can be generalized not only to other custom models fine-tuned from the same base model, but also to controllable generation using existing controllable tools. With the benefit of the decoupled cross-attention strategy, the image prompt can also work well with the text prompt to achieve multimodal image generation. The project page is available at \url{https://ip-adapter.github.io}.
\end{abstract}


\section{Introduction}
Image generation has made remarkable strides with the success of recent large text-to-image diffusion models like GLIDE~\cite{nichol2021glide}, DALL-E 2~\cite{ramesh2022hierarchical}, Imagen~\cite{saharia2022photorealistic}, Stable Diffusion (SD)~\cite{rombach2022high}, eDiff-I~\cite{balaji2022ediffi} and RAPHAEL~\cite{xue2023raphael}. Users can write text prompt to generate images with the powerful text-to-image diffusion models. But writing good text prompt to generate desired content is not easy, as complex prompt engineering~\cite{witteveen2022investigating} is often required. Moreover, text is not informative to express complex scenes or concepts, which can be a hindrance to content creation. Considering the above limitations of the text prompt, we may ask if there are other prompt types to generate images. A natural choice is to use the image prompt, since an image can express more content and details compared to text, just as often said: "an image is worth a thousand words". DALL-E 2\cite{ramesh2022hierarchical} makes the first attempt to support image prompt, the diffusion model is conditioned on image embedding rather than text embedding, and a prior model is required to achieve the text-to-image ability. However, most existing text-to-image diffusion models are conditioned on text to generate images, for example, the popular SD model is conditioned on the text features extracted from a frozen CLIP~\cite{radford2021learning} text encoder. Could image prompt be also supported on these text-to-image diffusion models? Our work attempts to enable the generative capability with image prompt for these text-to-image diffusion models in a simple manner.

Prior works, such as SD Image Variations\footnote{https://huggingface.co/lambdalabs/sd-image-variations-diffusers} and Stable unCLIP\footnote{https://huggingface.co/stabilityai/stable-diffusion-2-1-unclip}, have demonstrated the effectiveness of fine-tuning the text-conditioned diffusion models directly on image embedding to achieve image prompt capabilities. However, the disadvantages of this approach are obvious. First, it eliminates the original ability to generate images using text, and large 
computing resources are often required for such fine-tuning. Second, the fine-tuned models are typically not reusable, as the image prompt ability cannot be directly transferred to the other custom models derived from the same text-to-image base models. Moreover, the new models are often incompatible with existing structural control tools such as ControlNet~\cite{zhang2023adding}, which poses significant challenges for downstream applications. Due to the drawbacks of fine-tuning, some studies ~\cite{xu2023prompt} opt to replace the text encoder with an image encoder while avoiding fine-tuning the diffusion model. Although this method is effective and simple, it still has several drawbacks. At first, only the image prompt is supported, preventing users from simultaneously using text and image prompt to create images. Furthermore, merely fine-tuning the image encoder is often not sufficient to guarantee image quality, and could lead to generalization issues.

\renewcommand{\dblfloatpagefraction}{.8}
\begin{figure*}[t]
    \centering
    \includegraphics[width=0.95\textwidth]{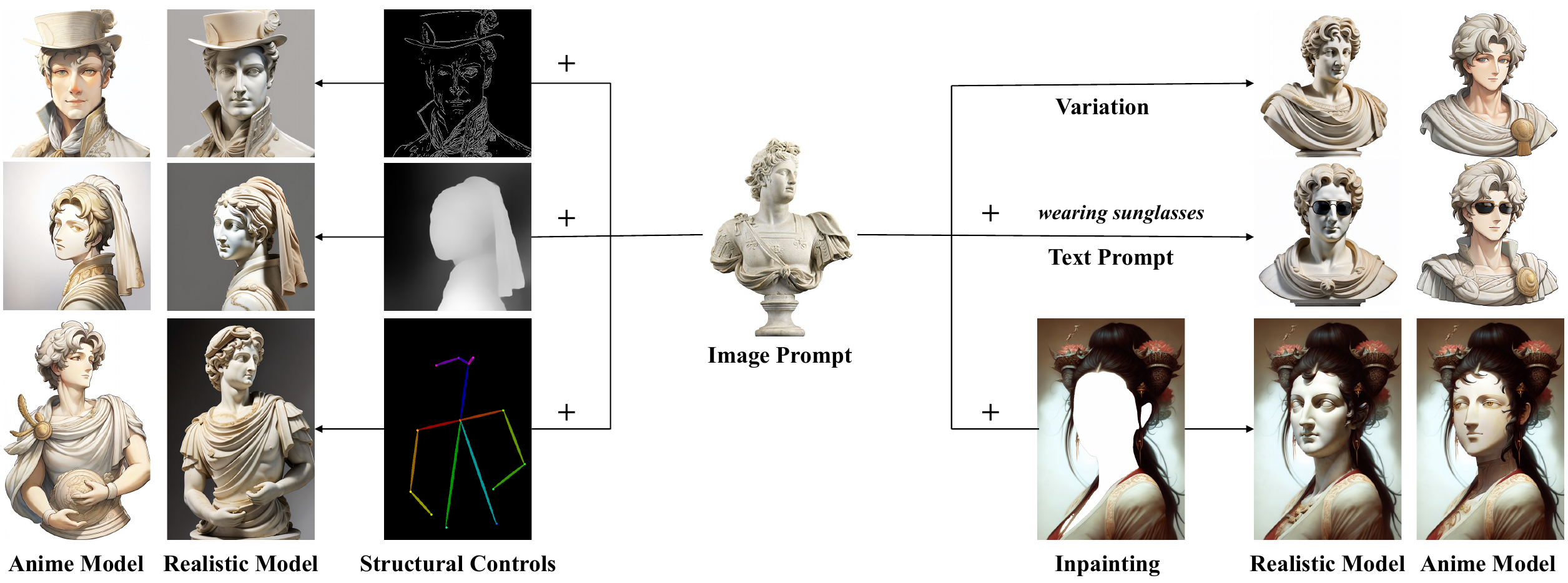}
     \setlength{\abovecaptionskip}{0.5cm}
     \vspace{-10pt}
    \caption{Various image synthesis with our proposed IP-Adapter applied on the pretrained text-to-image diffusion models with different styles. The examples on the right show the results of image variations, multimodal generation, and inpainting with image prompt, while the left examples show the results of controllable generation with image prompt and additional structural conditions.}
    \label{fig:result_overview}
    \vspace{-13pt}
\end{figure*}

In this study, we are curious about whether it is possible to achieve image prompt capability without modifying the original text-to-image models. Fortunately, previous works are encouraging. Recent advances in controllable image generation, such as ControlNet~\cite{zhang2023adding} and T2I-adapter~\cite{mou2023t2i}, have demonstrated that an additional network can be effectively plugged in the existing text-to-image diffusion models to guide the image generation. Most of the studies focus on image generation with additional structure control such as user-drawn sketch, depth map, semantic segmentation map, etc. Besides, image generation with style or content provided by reference image has also been achieved by simple adapters, such as the style adapter of T2I-adapter~\cite{mou2023t2i} and global controller of Uni-ControlNet~\cite{zhao2023uni}. To achieve this, image features extracted from CLIP image encoder are mapped to new features by a trainable network and then concatenated with text features. By replacing the original text features, the merged features are fed into the UNet of the diffusion model to guide image generation. These adapters can be seen as a way to have the ability to use image prompt, but the generated image is only partially faithful to the prompted image. The results are often worse than the fine-tuned image prompt models, let alone the model trained from scratch.

We argue that the main problem of the aforementioned methods lies in the cross-attention modules of text-to-image diffusion models. The key and value projection weights of the cross-attention layer in the pretrained diffusion model are trained to adapt the text features. Consequently, merging image features and text features into the cross-attention layer only accomplishes the alignment of image features to text features, but this potentially misses some image-specific information and eventually leads to only coarse-grained controllable generation (e.g., image style) with the reference image.

To this end, we propose a more effective image prompt adapter named IP-Adapter to avoid the shortcomings of the previous methods. Specifically, IP-Adapter adopts a decoupled cross-attention mechanism for text features and image features. For every cross-attention layer in the UNet of diffusion model, we add an additional cross-attention layer only for image features. In the training stage, only the parameters of the new cross-attention layers are trained, while the original UNet model remains frozen. Our proposed adapter is lightweight but very efficient: the generative performance of an IP-Adapter with only 22M parameters is comparable to a fully fine-tuned image prompt model from the text-to-image diffusion model. More importantly, our IP-Adapter exhibits excellent generalization capabilities and is compatible with text prompt. With our proposed IP-Adapter, various image generation tasks can be easily achieved, as illustrated in Figure \ref{fig:result_overview}.

To sum up, our contributions are as follows:

\begin{itemize}
    \item We present IP-Adapter, a lightweight image prompt adaptation method with the decoupled cross-attention strategy for existing text-to-image diffusion models. Quantitative and qualitative experimental results show that a small IP-Adapter with about 22M parameters is comparable or even better than the fully fine-tuned models for image prompt based generation.
    \item Our IP-Adapter is reusable and flexible. IP-Adapter trained on the base diffusion model can be generalized to other custom models fine-tuned from the same base diffusion model. Moreover, IP-Adapter is compatible with other controllable adapters such as ControlNet, allowing for an easy combination of image prompt with structure controls.
    \item Due to the decoupled cross-attention strategy, image prompt is compatible with text prompt to achieve multimodal image generation.
\end{itemize}

\section{Related Work}

We focus on designing an image prompt adapter for the existing text-to-image diffusion models. In this section, we review recent works on text-to-image diffusion models, as well as relevant studies on adapters for large models.

\subsection{Text-to-Image Diffusion Models}
Large text-to-image models are mainly divided into two categories: autoregressive models and diffusion models. Early works, such as DALLE~\cite{ramesh2021zero}, CogView~\cite{ding2021cogview,ding2022cogview2} and Make-A-Scene~\cite{gafni2022make}, are autoregressive models. For the autoregressive model, an image tokenizer like VQ-VAE~\cite{van2017neural} is used to convert an image to tokens, then an autoregressive transformer~\cite{vaswani2017attention} conditioned on text tokens is trained to predict image tokens. However, autoregressive models often require large parameters and computing resources to generate high-quality images, as seen in Parti~\cite{yu2022scaling}.

Recently, diffusion models (DMs)~\cite{sohl2015deep,song2020denoising,song2020score,dhariwal2021diffusion} has emerged as the new state-of-the-art model for text-to-image generation. As a pioneer, GLIDE uses a cascaded diffusion architecture with a 3.5B text-conditional diffusion model at $64\times64$ resolution and a 1.5B text-conditional upsampling diffusion model at $256\times256$ resolution. DALL-E 2 employs a diffusion model conditioned image embedding, and a prior model was trained to generate image embedding by giving a text prompt. DALL-E 2 not only supports text prompt for image generation but also image prompt. To enhance the text understanding, Imagen adopts  T5~\cite{raffel2020exploring}, a large transformer language model pretrained on text-only data, as the text encoder of diffusion model. Re-Imagen~\cite{chen2022re} uses retrieved information to improve the fidelity of generated images for rare or unseen entities. SD is built on the latent diffusion model~\cite{rombach2022high}, which operates on the latent space instead of pixel space, enabling SD to generate high-resolution images with only a diffusion model. To improve text alignment, eDiff-I was designed with an ensemble of text-to-image diffusion models, utilizing multiple conditions, including T5 text, CLIP text, and CLIP image embeddings. Versatile Diffusion~\cite{xu2022versatile} presents a unified multi-flow diffusion framework to support text-to-image, image-to-text, and variations within a single model. To achieve controllable image synthesis, Composer~\cite{huang2023composer} presents a joint fine-tuning strategy with various conditions on a pretrained diffusion model conditioned on image embedding. RAPHAEL introduces a mixture-of-experts (MoEs) strategy ~\cite{shazeer2017outrageously,fedus2022switch} into the text-conditional image diffusion model to enhance image quality and aesthetic appeal.

An attractive feature of DALL-E 2 is that it can also use image prompt to generate image variations. Hence, there are also some works to explore to support image prompt for the text-to-image diffusion models conditioned only on text. SD Image Variations model is fine-tuned from a modified SD model where the text features are replaced with the image embedding from CLIP image encoder. Stable unCLIP is also a fine-tuned model on SD, in which the image embedding is added to the time embedding. Although the fine-tuning model can successfully use image prompt to generate images, it often requires a relatively large training cost, and it fails to be compatible with existing tools, e.g., ControlNet~\cite{zhang2023adding}.

\subsection{Adapters for Large Models}

As fine-tuning large pre-trained models is inefficient, an alternative approach is using adapters, which add a few trainable parameters but freeze the original model. Adapters have been used in the field of NLP for a long time~\cite{houlsby2019parameter}. Recently, adapters have been utilized to achieve vision-language understanding for large language models~\cite{li2023blip,zhu2023minigpt,zhang2023llama,gao2023llama,zeng2023matters}.

With the popularity of recent text-to-image models, adapters have also been used to provide additional control for the generation of text-to-image models. ControlNet~\cite{zhang2023adding} first proves that an adapter could be trained with a pretrained text-to-image diffusion model to learn task-specific input conditions, e.g., canny edge. Almost concurrently, T2I-adapter~\cite{mou2023t2i} employs a simple and lightweight adapter to achieve
fine-grained control in the color and structure of the generated images. To reduce the fine-tuning cost, Uni-ControlNet~\cite{zhao2023uni} presents a multi-scale condition injection strategy to learn an adapter for various local controls.

Apart from the adapters for structural control, there are also works for the controllable generation conditioned on the content and style of the provided image. ControlNet Shuffle \footnote{https://github.com/lllyasviel/ControlNet-v1-1-nightly} trained to recompose images, can be used to guide the generation by a user-provided image. Moreover, ControlNet Reference-only\footnote{https://github.com/Mikubill/sd-webui-controlnet} was presented to achieve image variants on SD model through simple feature injection without training. In the updated version of T2I-adapter, a style adapter is designed to control the style of generated images using a reference image by appending image features extracted from the CLIP image encoder to text features. The global control adapter of Uni-ControlNet also projects the image embedding from CLIP image encoder into condition embeddings by a small network and concatenates them with the original text embeddings, and it is used to guide the generation with the style and content of reference image. SeeCoder~\cite{xu2023prompt} presents a semantic context encoder to replace the original text encoder to generate image variants.

Although the aforementioned adapters are lightweight, their performance is hardly comparable to that of the fine-tuned image prompt models, let alone one trained from scratch. In this study, we introduce a decoupled cross-attention mechanism to achieve a more effective image prompt adapter. The proposed adapter remains simple and small but outperforms previous adapter methods, and is even comparable to fine-tuned models.
\renewcommand{\dblfloatpagefraction}{.9}
\begin{figure}[t]
    \centering
    \includegraphics[width=0.9\textwidth]{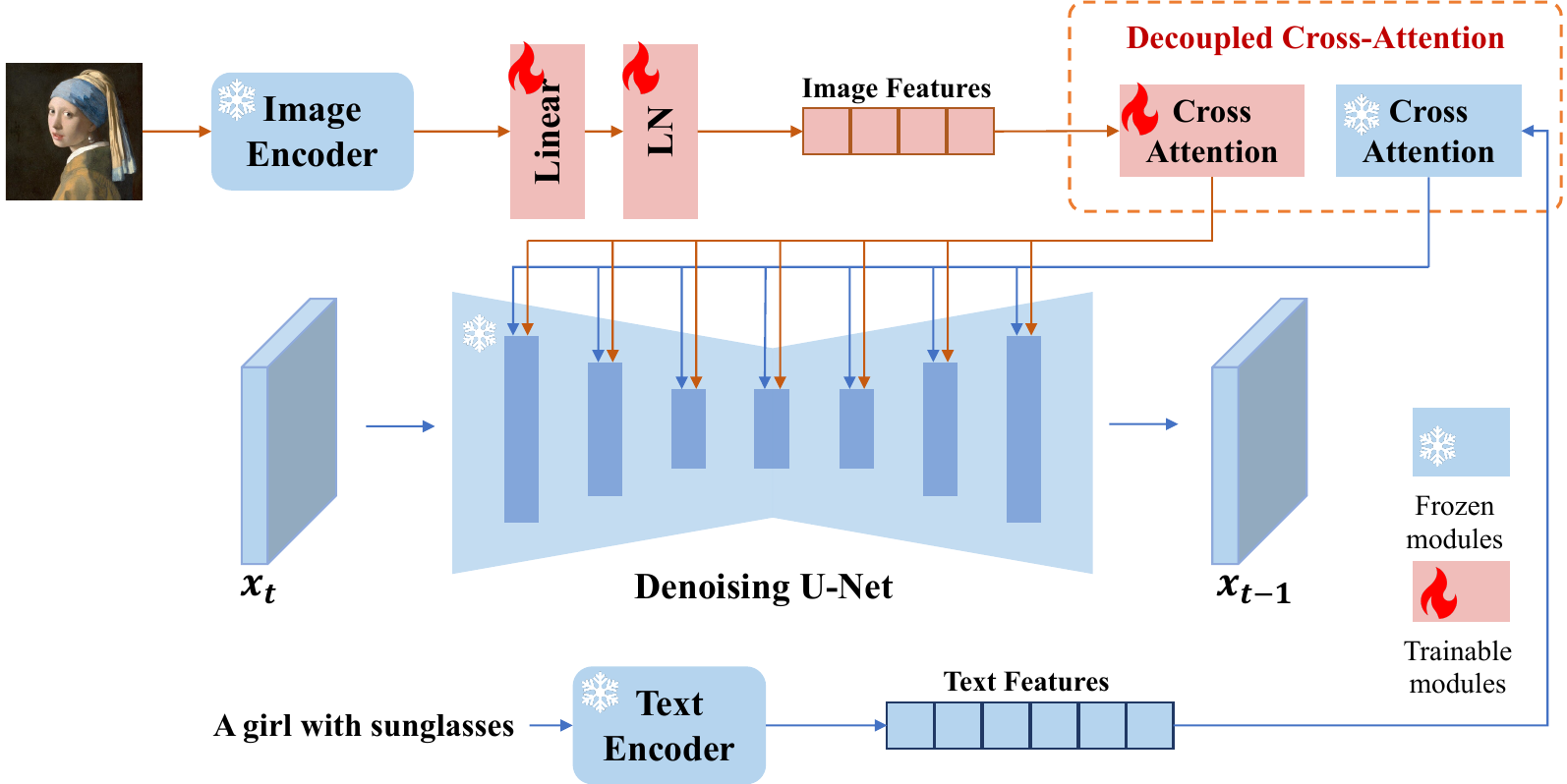}
     \setlength{\abovecaptionskip}{0.5cm}
     \vspace{-10pt}
    \caption{The overall architecture of our proposed IP-Adapter with decoupled cross-attention strategy. Only the newly added modules (in red color) are trained while the pretrained text-to-image model is frozen.
}
    \label{fig:architecture}
    \vspace{-13pt}
\end{figure}
\section{Method}
In this section, we first introduce some preliminaries about text-to-image diffusion models. Then, we depict in detail the motivation and the design of the proposed IP-Adapter.
\subsection{Prelimiaries}
Diffusion models are a class of generative models that comprise two processes: a diffusion process (also known as the forward process), which gradually adds Gaussian noise to the data using a fixed Markov chain of $T$ steps, and a denoising process that generates samples from Gaussian noise with a learnable model. Diffusion models can also be conditioned on other inputs, such as text in the case of text-to-image diffusion models. Typically, the training objective of a diffusion model, denoted as $\boldsymbol{\epsilon}_{\theta}$, which predicts noise, is defined as a simplified variant of the variational bound:
\begin{equation}
L_{\text{simple}}=\mathbb{E}_{\boldsymbol{x}_{0},\boldsymbol{\epsilon}\sim \mathcal{N}(\mathbf{0}, \mathbf{I}), \boldsymbol{c}, t} \| \boldsymbol{\epsilon}- \boldsymbol{\epsilon}_\theta\big(\boldsymbol{x}_t, \boldsymbol{c}, t\big)\|^2,
\end{equation}
where $\boldsymbol{x}_{0}$ represents the real data with an additional condition $\boldsymbol{c}$, $t\in [0, T]$ denotes the time step of diffusion process, $\boldsymbol{x}_t = \alpha_t\boldsymbol{x}_0+\sigma_t\boldsymbol{\epsilon}$ is the noisy data at $t$ step, and $\alpha_t$, $\sigma_t$ are predefined functions of $t$ that determine the diffusion process. Once the model $\boldsymbol{\epsilon}_{\theta}$ is trained, images can be generated from random noise in an iterative manner. Generally, fast samplers such as DDIM~\cite{song2020denoising}, PNDM~\cite{liu2022pseudo} and DPM-Solver~\cite{lu2022dpm,lu2022dpm++}, are adopted in the inference stage to accelerate the generation process.

For the conditional diffusion models, classifier guidance~\cite{dhariwal2021diffusion} is a straightforward technique used to balance image fidelity and sample diversity by utilizing gradients from a separately trained classifier. To eliminate the need for training a classifier independently, classifier-free guidance~\cite{ho2022classifier} is often employed as an alternative method. In this approach, the conditional and unconditional diffusion models are jointly trained by randomly dropping $\boldsymbol{c}$ during training. In the sampling stage, the predicted noise is calculated based on the prediction of both the conditional model $\boldsymbol{\epsilon}_{\theta}(\boldsymbol{x}_t, \boldsymbol{c}, t)$ and unconditional model $\boldsymbol{\epsilon}_{\theta}(\boldsymbol{x}_t, t)$:
\begin{equation}
\hat{\boldsymbol{\epsilon}}_{\theta}(\boldsymbol{x}_t, \boldsymbol{c}, t) = w\boldsymbol{\epsilon}_{\theta}(\boldsymbol{x}_t, \boldsymbol{c}, t)+(1-w)\boldsymbol{\epsilon}_{\theta}(\boldsymbol{x}_t, t),
\end{equation}
here, $w$, often named guidance scale or guidance weight, is a scalar value that adjusts the alignment with condition $\boldsymbol{c}$. For text-to-image diffusion models, classifier-free guidance plays a crucial role in enhancing the image-text alignment of generated samples.

In our study, we utilize the open-source SD model as our example base model to implement the IP-Adapter. SD is a latent diffusion model conditioned on text features extracted from a frozen CLIP text encoder. The architecture of the diffusion model is based on a UNet~\cite{ronneberger2015u} with attention layers. Compared to pixel-based diffusion models like Imagen, SD is more efficient since it is constructed on the latent space from a pretrained auto-encoder model.
\subsection{Image Prompt Adapter}
In this paper, the image prompt adapter is designed to enable a pretrained text-to-image diffusion model to generate images with image prompt. As mentioned in previous sections, current adapters struggle to match the performance of fine-tuned image prompt models or the models trained from scratch. The major reason is that the image features cannot be effectively embedded in the pretrained model. Most methods simply feed concatenated features into the frozen cross-attention layers, preventing the diffusion model from capturing fine-grained features from the image prompt. To address this issue, we present a decoupled cross-attention strategy, in which the image features are embedded by newly added cross-attention layers. The overall architecture of our proposed IP-Adapter is demonstrated in Figure~\ref{fig:architecture}. The proposed IP-Adapter consists of two parts: an image encoder to extract image features from image prompt, and adapted modules with decoupled cross-attention to embed image features into the pretrained text-to-image diffusion model. 
\subsubsection{Image Encoder}
Following most of the methods, we use a pretrained CLIP image encoder model to extract image features from the image prompt. The CLIP model is a multimodal model trained by contrastive learning on a large dataset containing image-text pairs. We utilize the global image embedding from the CLIP image encoder, which is well-aligned with image captions and can represent the rich content and style of the image. In the training stage, the CLIP image encoder is frozen.

To effectively decompose the global image embedding, we use a small trainable projection network to project the image embedding into a sequence of features with length $N$ (we use $N$ = 4 in this study), the dimension of the image features is the same as the dimension of the text features in the pretrained diffusion model. The projection network we used in this study consists of a linear layer and a Layer Normalization~\cite{ba2016layer}. 
\subsubsection{Decoupled Cross-Attention}
The image features are integrated into the pretrained UNet model by the adapted modules with decoupled cross-attention. In the original SD model, the text features from the CLIP text encoder are plugged into the UNet model by feeding into the cross-attention layers. Given the query features $\mathbf{Z}$ and the text features $\boldsymbol{c}_{t}$, the output of cross-attention $\mathbf{Z}'$ can be defined by the following equation:
\begin{equation}
\begin{split}
\mathbf{Z}'=\text{Attention}(\mathbf{Q},\mathbf{K},\mathbf{V}) = \text{Softmax}(\frac{\mathbf{Q}\mathbf{K}^{\top}}{\sqrt{d}})\mathbf{V},
\\
\end{split}
\end{equation}
where $\mathbf{Q}=\mathbf{Z}\mathbf{W}_q$, $\mathbf{K}=\boldsymbol{c}_{t}\mathbf{W}_k$, $\mathbf{V}=\boldsymbol{c}_{t}\mathbf{W}_v$ are the query, key, and values matrices of the attention operation respectively, and $\mathbf{W}_q$, $\mathbf{W}_k$, $\mathbf{W}_v$ are the weight matrices of the trainable linear projection layers.

A straightforward method to insert image features is to concatenate image features and text features and then feed them into the cross-attention layers. However, we found this approach to be insufficiently effective. Instead, we propose a decoupled cross-attention mechanism where the cross-attention layers for text features and image features are separate. To be specific, we add a new cross-attention layer for each cross-attention layer in the original UNet model to insert image features. Given the image features $\boldsymbol{c}_{i}$, the output of new cross-attention $\mathbf{Z}''$ is computed as follows:
\begin{equation}
    \begin{split}
    \mathbf{Z}''=\text{Attention}(\mathbf{Q},\mathbf{K}',\mathbf{V}') = \text{Softmax}(\frac{\mathbf{Q}(\mathbf{K}')^{\top}}{\sqrt{d}})\mathbf{V}',\\
    \end{split}
\end{equation}
where, $\mathbf{Q}=\mathbf{Z}\mathbf{W}_q$, $\mathbf{K}'=\boldsymbol{c}_{i}\mathbf{W}'_k$ and $\mathbf{V}'=\boldsymbol{c}_{i}\mathbf{W}'_v$ are the query, key, and values matrices from the image features. $\mathbf{W}'_k$ and $\mathbf{W}'_v$ are the corresponding weight matrices. It should be noted that we use the same query for image cross-attention as for text cross-attention. Consequently, we only need add two paramemters $\mathbf{W}'_k$, $\mathbf{W}'_v$ for each cross-attention layer. In order to speed up the convergence, $\mathbf{W}'_k$ and $\mathbf{W}'_v$ are initialized from $\mathbf{W}_k$ and $\mathbf{W}_v$.
Then, we simply add the output of image cross-attention to the output of text cross-attention. Hence, the final formulation of the decoupled cross-attention is defined as follows:
\begin{equation}
\begin{split}
\mathbf{Z}^{new}=\text{Softmax}(\frac{\mathbf{Q}\mathbf{K}^{\top}}{\sqrt{d}})\mathbf{V}+\text{Softmax}(\frac{\mathbf{Q}(\mathbf{K}')^{\top}}{\sqrt{d}})\mathbf{V}'\\
 \text{where}\ \mathbf{Q}=\mathbf{Z}\mathbf{W}_q, \mathbf{K}=\boldsymbol{c}_{t}\mathbf{W}_k, \mathbf{V}=\boldsymbol{c}_{t}\mathbf{W}_v,
 \mathbf{K}'=\boldsymbol{c}_{i}\mathbf{W}'_k, \mathbf{V}'=\boldsymbol{c}_{i}\mathbf{W}'_v
\end{split}
\end{equation}
Sine we freeze the original UNet model, only the $\mathbf{W}'_k$ and $\mathbf{W}'_v$ are trainable in the above decoupled cross-attention. 
\subsubsection{Training and Inference}
During training, we only optimize the IP-Adapter while keeping the parameters of the pretrained diffusion model fixed. The IP-Adapter is also trained on the dataset with image-text pairs\footnote{Note that it is also possible to train the model without text prompt since using image prompt only is informative to guide the final generation.}, using the same training objective as original SD:
\begin{equation}
L_{\text{simple}}=\mathbb{E}_{\boldsymbol{x}_{0},\boldsymbol{\epsilon}, \boldsymbol{c}_{t}, \boldsymbol{c}_{i}, t} \| \boldsymbol{\epsilon}- \boldsymbol{\epsilon}_\theta\big(\boldsymbol{x}_t, \boldsymbol{c}_{t}, \boldsymbol{c}_{i}, t\big)\|^2.
\end{equation}
We also randomly drop image conditions in the training stage to enable classifier-free guidance in the inference stage:
\begin{equation}
\begin{split}
\hat{\boldsymbol{\epsilon}}_{\theta}(\boldsymbol{x}_t, \boldsymbol{c}_{t}, \boldsymbol{c}_{i}, t) = w\boldsymbol{\epsilon}_{\theta}(\boldsymbol{x}_t, \boldsymbol{c}_{t}, \boldsymbol{c}_{i}, t)+(1-w)\boldsymbol{\epsilon}_{\theta}(\boldsymbol{x}_t, t)
\end{split}
\end{equation}
Here, we simply zero out the CLIP image embedding if the image condition is dropped. 

As the text cross-attention and image cross-attention are detached, we can also adjust the weight of the image condition in the inference stage:
\begin{equation}
\mathbf{Z}^{new}=\text{Attention}(\mathbf{Q},\mathbf{K},\mathbf{V}) + \lambda\cdot\text{Attention}(\mathbf{Q},\mathbf{K}',\mathbf{V}')
\end{equation}
where $\lambda$ is weight factor, and the model becomes the original text-to-image diffusion model if $\lambda=0$. 

\section{Experiments}
\subsection{Experimental Setup}
\subsubsection{Training Data}
To train the IP-Adapter, we build a multimodal dataset including  about 10 million text-image
pairs from two open source datasets - LAION-2B~\cite{schuhmann2022laion} and COYO-700M~\cite{kakaobrain2022coyo-700m}.
\subsubsection{Implementation Details}
Our experiments are based on SD v1.5\footnote{https://huggingface.co/runwayml/stable-diffusion-v1-5}, and we use OpenCLIP ViT-H/14~\cite{ilharco2021openclip} as the image encoder. There are 16 cross-attention layers in SD model, and we add a new image cross-attention layer for each of these layers. The total trainable parameters of our IP-Adapter including a projection network and adapted modules, amount to about 22M, making the IP-Adapter quite lightweight. We implement our IP-Adapter with HuggingFace diffusers library~\cite{von-platen-etal-2022-diffusers} and employ DeepSpeed ZeRO-2~\cite{ramesh2021zero} for fast training. IP-Adapter is trained on a single machine with 8 V100 GPUs for 1M steps with a batch size of 8 per GPU. We use the AdamW optimizer~\cite{loshchilov2017decoupled} with a fixed learning rate of 0.0001 and weight decay of 0.01. During training, we resize the shortest side of the image to 512 and then center crop the image with $512\times512$ resolution. To enable classifier-free guidance, we use a probability of 0.05 to drop text and image individually, and a probability of 0.05 to drop text and image simultaneously. In the inference stage, we adopt DDIM sampler with 50 steps, and set the guidance scale to 7.5. When only using image prompt, we set the text prompt to empty and $\lambda=1.0$.

\begin{table}[ht]
    \centering
    \footnotesize
    \caption{Quantitative comparison of the proposed IP-Adapter with other methods on COCO validation set. The best results are in \textbf{bold}.}
    \resizebox{\linewidth}{!}{
    \begin{tabular}{ccccccc}
\toprule
  Method & \makecell[c]{Reusable to \\custom models} & \makecell[c]{Compatible with \\controllable tools} & \makecell[c]{Multimodal \\prompts} & \makecell[c]{Trainable \\parameters} & CLIP-T $\uparrow$ & CLIP-I $\uparrow$\\
\midrule
\emph{Training from scratch} \\
\midrule

Open unCLIP & \XSolidBrush & \XSolidBrush & \XSolidBrush & 893M &\textbf{0.608} &\textbf{0.858} \\

Kandinsky-2-1 & \XSolidBrush & \XSolidBrush & \XSolidBrush & 1229M &0.599 &0.855 \\

Versatile Diffusion & \XSolidBrush & \XSolidBrush & \Checkmark & 860M& 0.587 & 0.830\\
\midrule
\emph{ Fine-tunining from text-to-image model } \\
\midrule
SD Image Variations &\XSolidBrush & \XSolidBrush & \XSolidBrush &860M &0.548 &0.760 \\
SD unCLIP &\XSolidBrush & \XSolidBrush & \XSolidBrush & 870M & 0.584 & 0.810 \\
\midrule
\emph{Adapters} \\
\midrule

Uni-ControlNet (Global Control) & \Checkmark & \Checkmark & \Checkmark &47M & 0.506 & 0.736 \\

T2I-Adapter (Style) & \Checkmark & \Checkmark & \Checkmark & 39M & 0.485 & 0.648 \\

ControlNet Shuffle & \Checkmark & \Checkmark & \Checkmark & 361M & 0.421 & 0.616 \\

\textbf{IP-Adapter} & \Checkmark & \Checkmark & \Checkmark & 22M & \textbf{0.588} & \textbf{0.828} \\
\bottomrule
    \end{tabular}}
    
    \label{tab:cmp_result}
\end{table}

\renewcommand{\dblfloatpagefraction}{.8}
\begin{figure*}[htbp]
    \centering
    \includegraphics[width=1.0\textwidth]{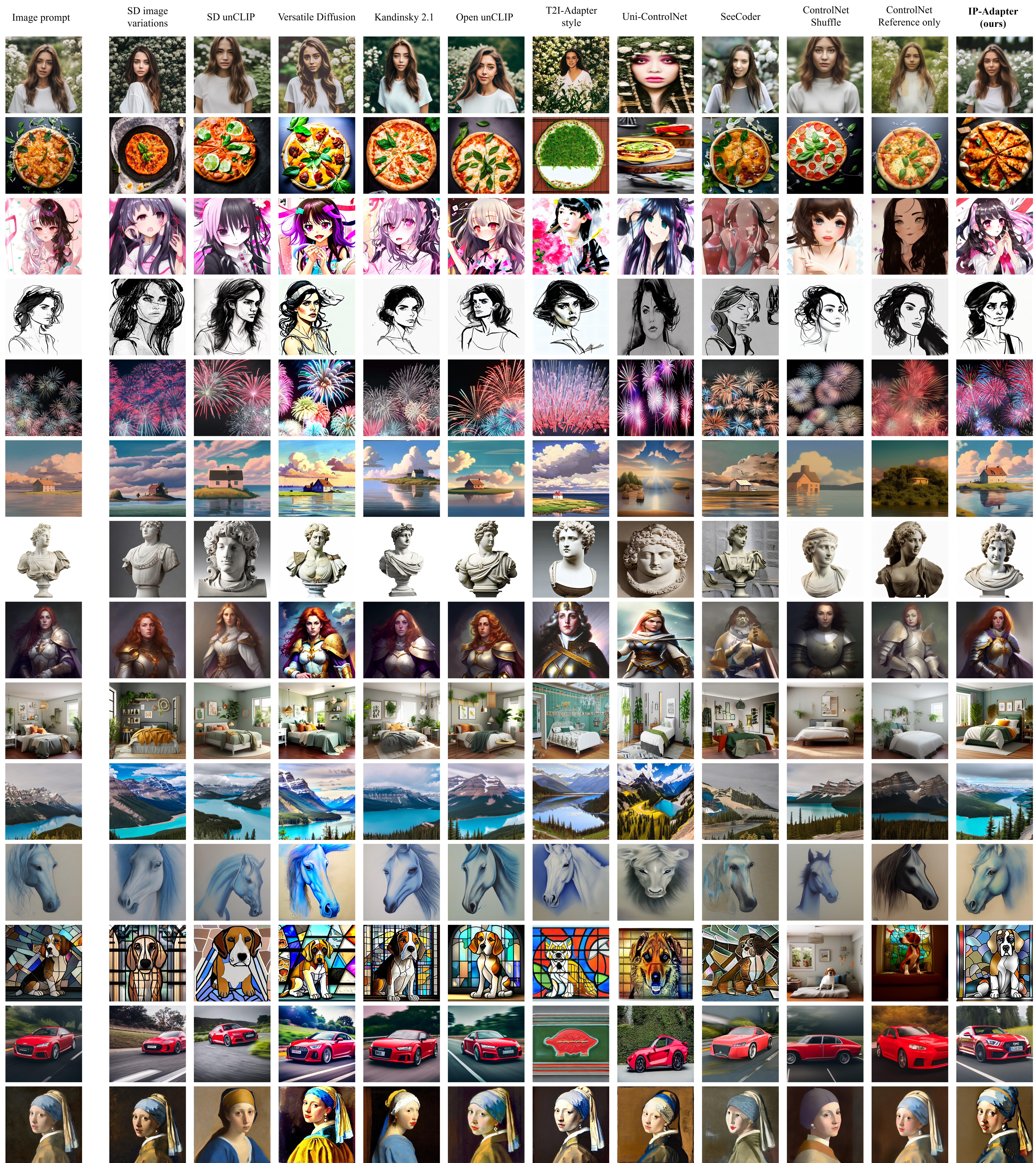}
     \vspace{-10pt}
    \caption{The visual comparison of our proposed IP-Adapter with other methods conditioned on different kinds and styles of images.
}
    \label{fig:result1}
    \vspace{-13pt}
\end{figure*}

\renewcommand{\dblfloatpagefraction}{.8}
\begin{figure*}[!h]
    \centering
    \setlength{\abovecaptionskip}{0.5cm}
    \includegraphics[width=0.95\textwidth]{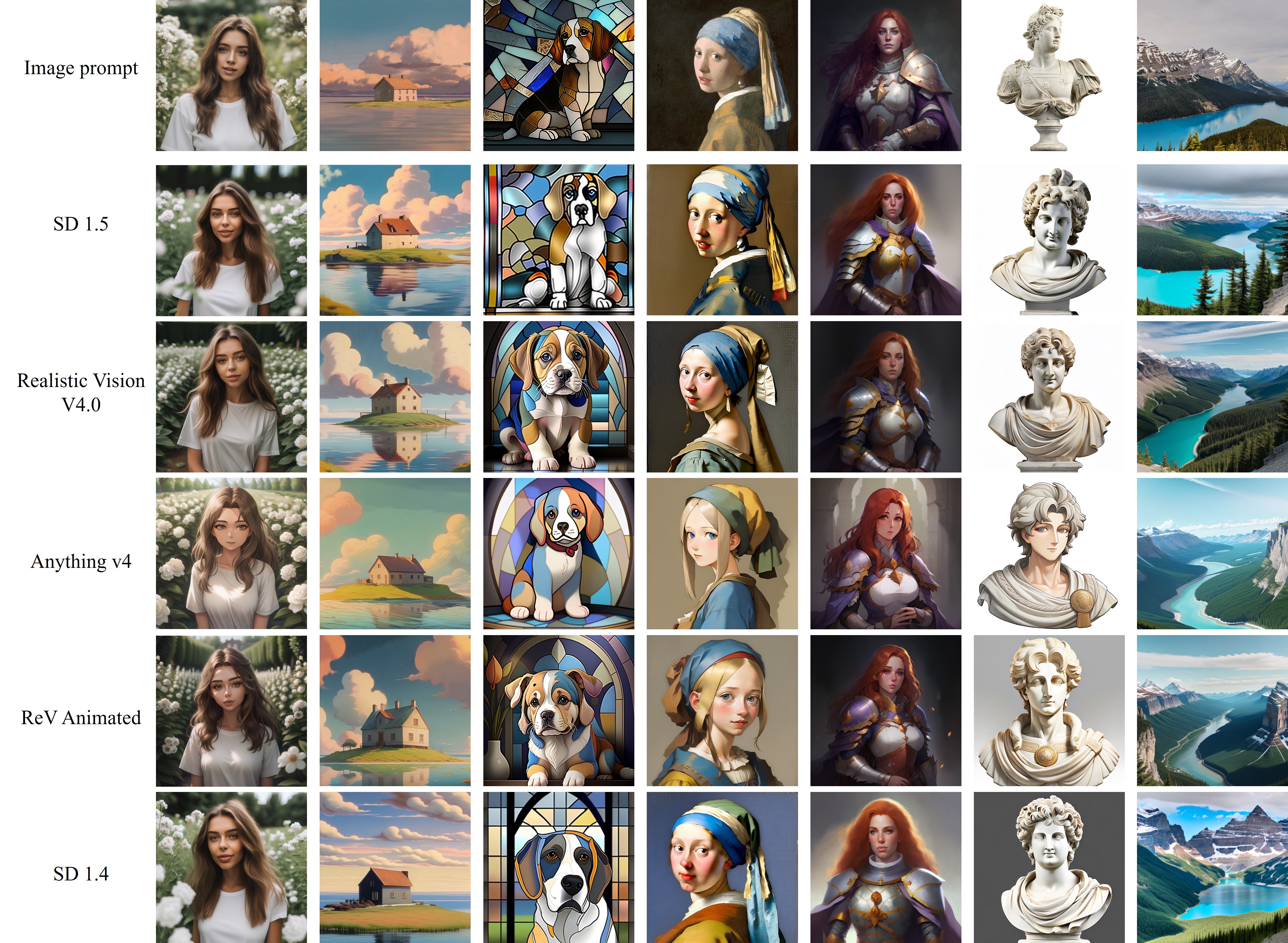}
     \vspace{-10pt}
    \caption{The generated images of different diffusion models with our proposed IP-Adapter. The IP-Adapter is only trained once.
}
    \label{fig:result2}
    \vspace{-13pt}
\end{figure*}

\renewcommand{\dblfloatpagefraction}{.8}
\begin{figure*}[htbp]
    \centering
    \includegraphics[width=1.0\textwidth]{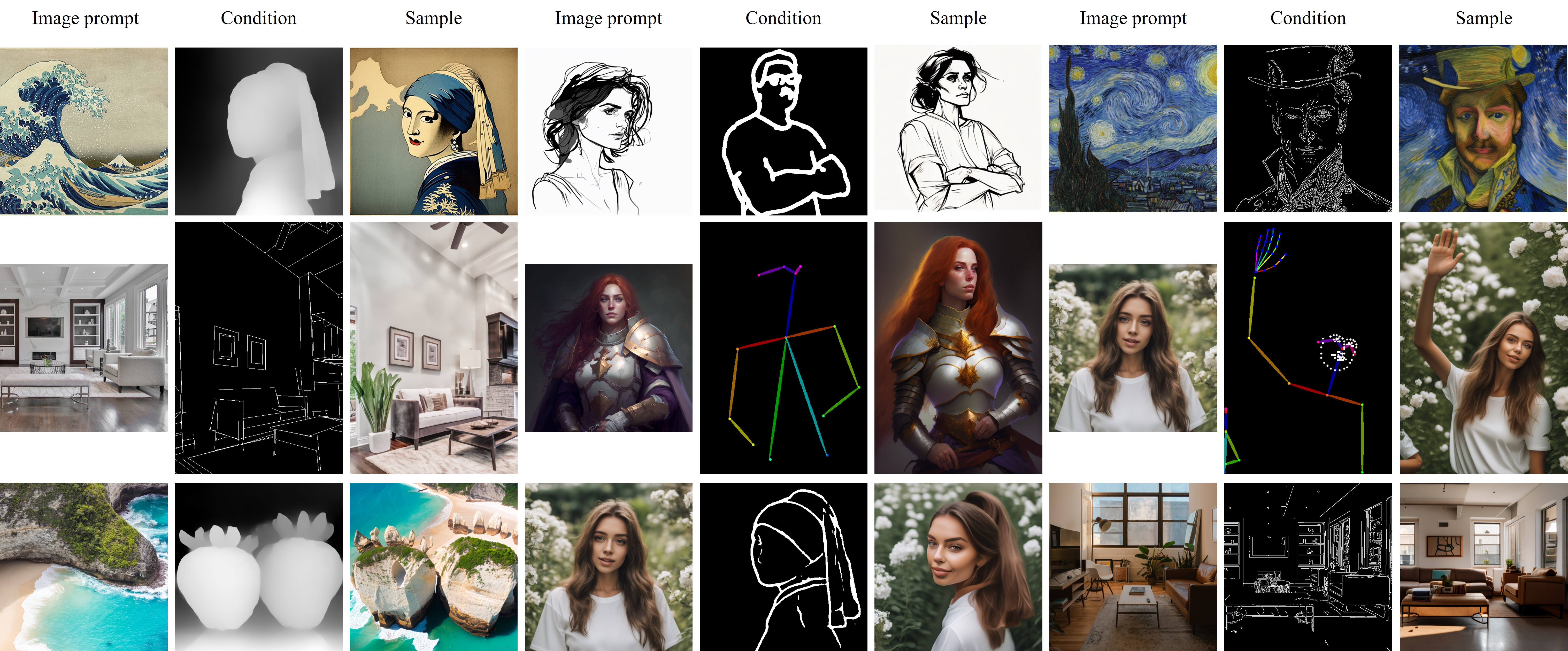}
     \vspace{-10pt}
    \caption{Visualization of generated samples with image prompt and additional structural conditions. Note that we don't need fine-tune the IP-Adapter.
}
    \label{fig:result3}
    \vspace{-13pt}
\end{figure*}
\begin{figure*}[!ht]
    \centering
    \includegraphics[width=1.0\textwidth]{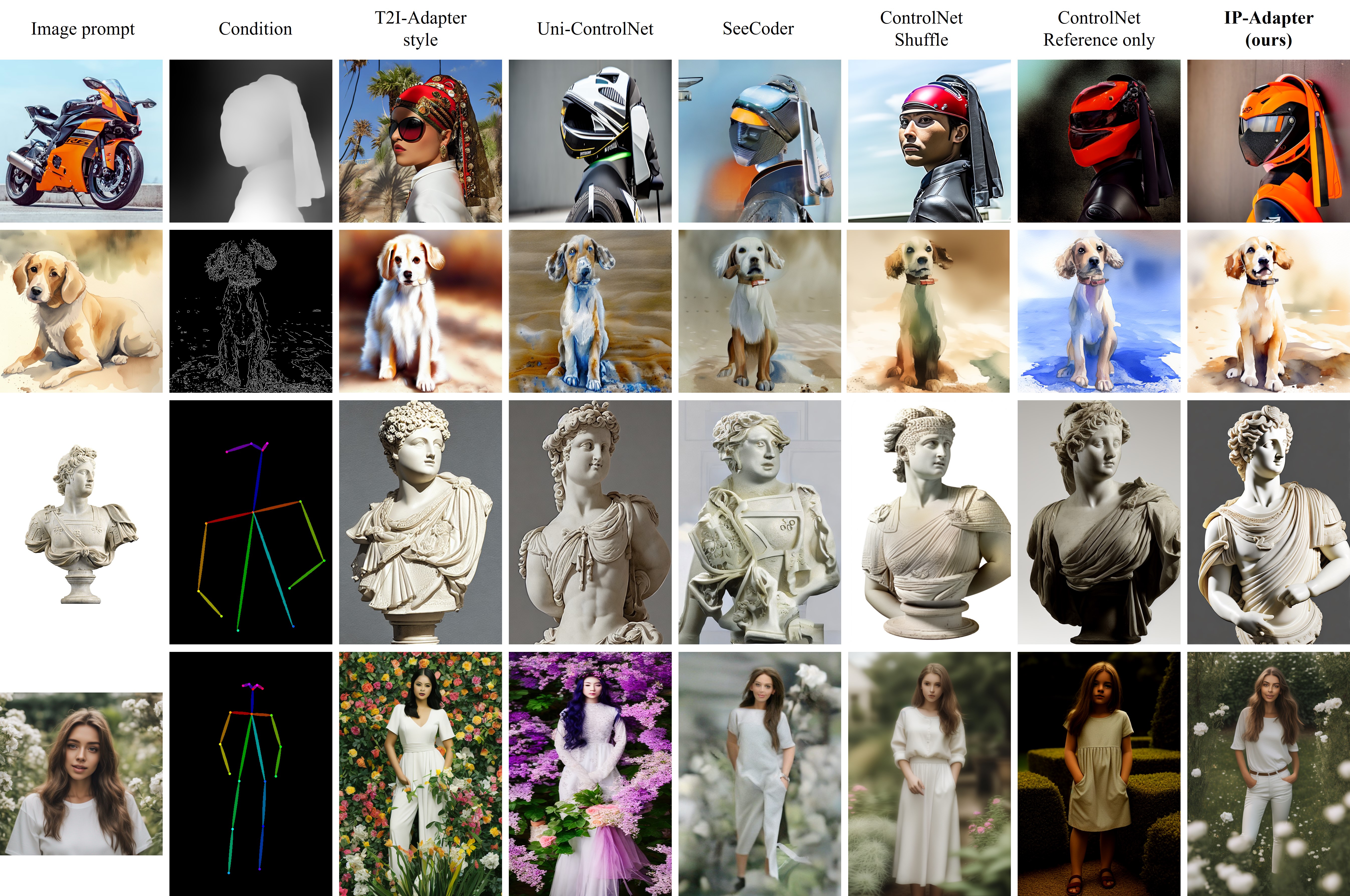}
     \vspace{-10pt}
    \caption{Comparison of our IP-Adapter with other methods on different structural conditions.
}
    \label{fig:result4}
    \vspace{-13pt}
\end{figure*}
\renewcommand{\dblfloatpagefraction}{.8}
\begin{figure*}[!h]
    \centering
    \setlength{\abovecaptionskip}{0.5cm}
    \setlength{\belowcaptionskip}{0.25cm}
    \includegraphics[width=0.85\textwidth]{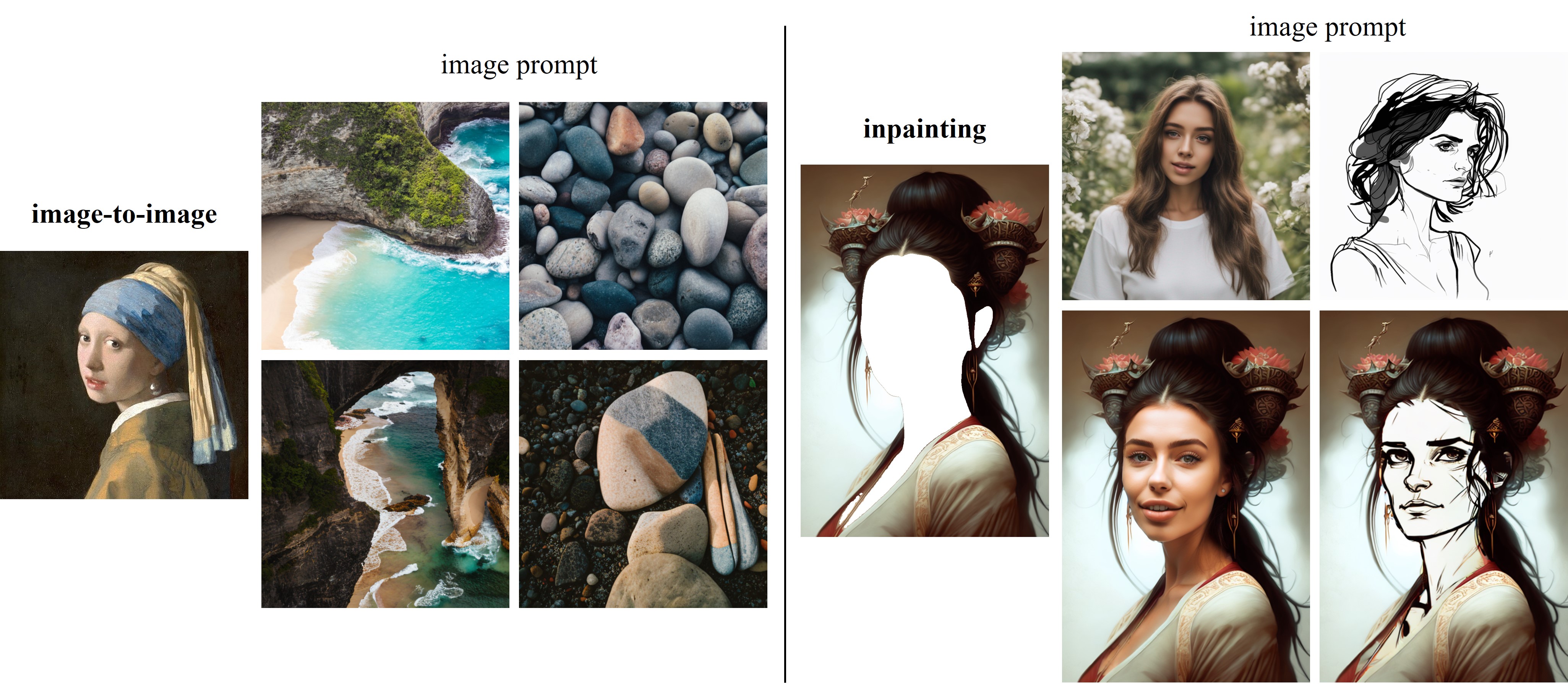}
     \vspace{-10pt}
    \caption{Examples of image-to-image and inpainting with image prompt by our IP-Adapter.
}
    \label{fig:result5}
    \vspace{-13pt}
\end{figure*}
\subsection{Comparison with Existing Methods}
To demonstrate the effectiveness of our method, we compare our IP-Adapter with other existing methods on generation with image prompt. We select three types of methods: training from scratch, fine-tuning from text-to-image model, and adapters. For the method trained from scratch, we select 3 open source models: open unCLIP\footnote{https://github.com/kakaobrain/karlo} which is a reproduction of DALL-E 2, Kandinsky-2-1~\footnote{https://github.com/ai-forever/Kandinsky-2} which is a mixture of DALL-E 2 and latent diffusion, and Versatile Diffusion~\cite{xu2022versatile}. For the fine-tuned models, we choose SD Image Variations and SD unCLIP. For the adapters, we compare our IP-Adapter with the style-adapter of T2I-Adapter, the global controller of Uni-ControlNet, ControlNet Shuffle, ControlNet Reference-only and SeeCoder.

\subsubsection{Quantitative Comparison}
We use the validation set of COCO2017~\cite{lin2014microsoft} containing 5,000 images with captions for quantitative evaluation. For a fair comparison, we generate 4 images conditioned on the image prompt for each sample in the dataset, resulting in total 20,000 generated images for each method. We use two metrics to evaluate the alignment with the image condition: 
\begin{itemize}
\item CLIP-I: the similarity in CLIP image embedding of generated images with
the image prompt.
\item CLIP-T: the CLIPScore~\cite{hessel2021clipscore} of the generated images with
captions of the image prompts.
\end{itemize}
We calculate the average value of the two metrics on all generated images with CLIP ViT-L/14\footnote{https://huggingface.co/openai/clip-vit-large-patch14} model. As the open source SeeCoder is used with additional structural controls and ControlNet Reference-only is released under the web framework, we only conduct qualitative evaluations. The comparison results are shown in Table \ref{tab:cmp_result}. As we observe, our method is much better than other adapters, and is also comparable or even better than the fine-tuned model with only 22M parameters.
\subsubsection{Qualitative Comparison}
We also select various kinds and styles of images to qualitatively evaluate our method. For privacy reasons, the images with real face are synthetic. For SeeCoder, we also use the scribble control with ControlNet to generate images. For ControlNet Reference-only, we also input the captions generated with BLIP caption model~\cite{li2022blip}. For each image prompt, we random generate 4 samples and select the best one for each method to ensure fairness. As we can see in Figure \ref{fig:result1}, the proposed IP-Adapter is mostly better than other adapters both in image quality and alignment with the reference image. Moreover, our method is slightly better than the fine-tuned models, and also comparable to the models trained from scratch in most cases.

In conclusion, the proposed IP-Adapter is lightweight and effective method to achieve the generative capability with image prompt for the pretrained text-to-image diffusion models.

\subsection{More Results}
Although the proposed IP-Adapter is designed to achieve the generation with image prompt, its robust generalization capabilities allow for a broader range of applications. As shown in Table \ref{tab:cmp_result}, our IP-Adapter is not only reusable to custom models, but also compatible with existing controllable tools and text prompt. In this part, we show more results that our adapter can generate.
\subsubsection{Generalizable to Custom Models}
As we freeze the original diffusion model in the training stage, the IP-Adapter can also be generalizable to the custom models fine-tuned from SD v1.5 like other adapters (e.g., ControlNet). In other words, once IP-Adapter is trained, it can be directly reusable on custom models fine-tuned from the same base model. To validate this, we select three community models from HuggingFace model library\footnote{https://huggingface.co/models}: Realistic Vision V4.0, Anything v4, and ReV Animated. These models are all fine-tuned from SD v1.5. As shown in Figure \ref{fig:result2}, our IP-Adapter works well on these community models. Furthermore, the generated images can mix the style of the community models, for example, we can generate anime-style images when using the anime-style model Anything v4. Interestingly, our adapter can be directly applied to SD v1.4, as SD v1.5 is trained with more steps based on SD v1.4.

\renewcommand{\dblfloatpagefraction}{.8}
\begin{figure*}[!t]
    \centering
    \setlength{\abovecaptionskip}{0.5cm}
    \includegraphics[width=0.95\textwidth]{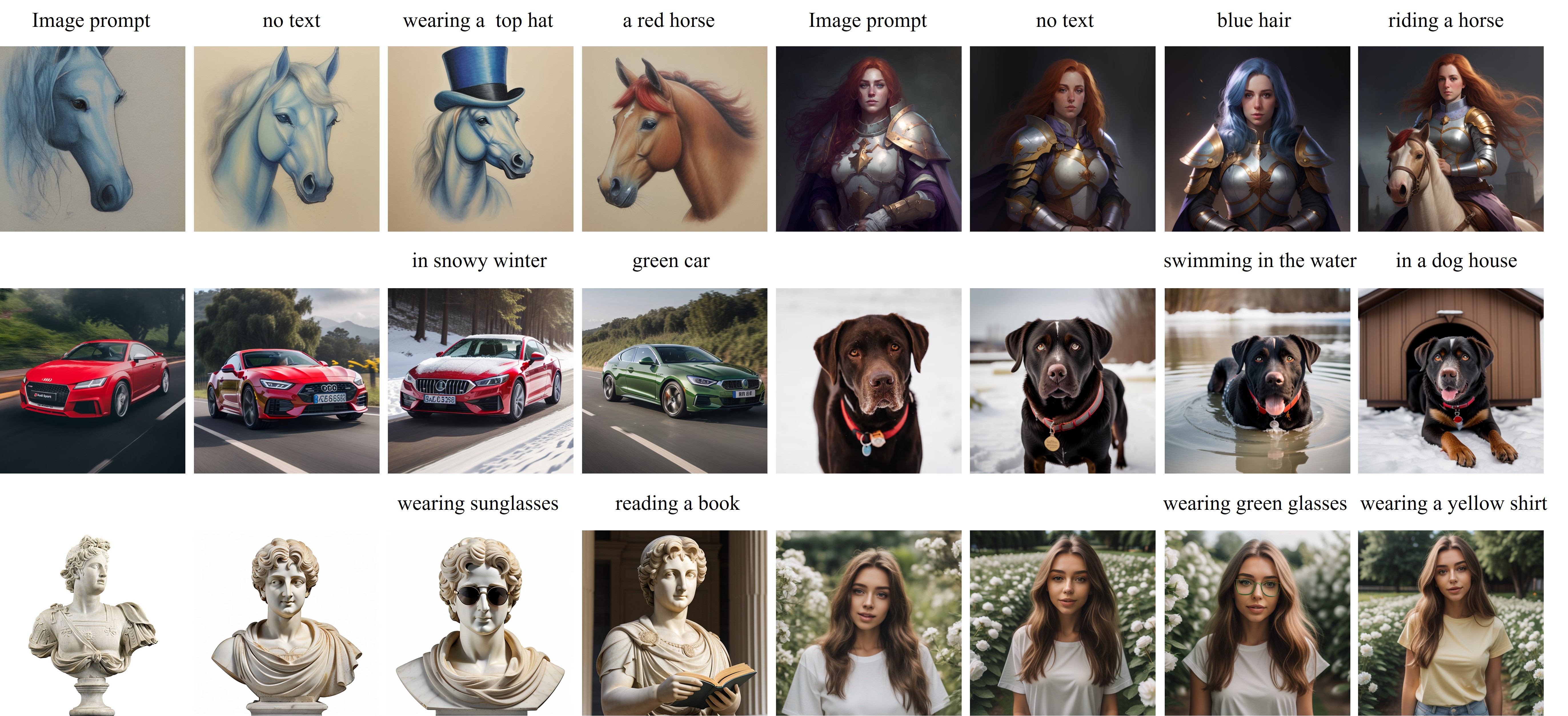}
     \vspace{-10pt}
    \caption{Generated examples of our IP-Adapter with multimodal prompts.}
    \label{fig:result6}
    \vspace{-13pt}
\end{figure*}

\renewcommand{\dblfloatpagefraction}{1.0}
\begin{figure*}[!h]
    \centering
    \setlength{\abovecaptionskip}{0.5cm}
    \setlength{\belowcaptionskip}{0.25cm}
    \includegraphics[width=1.0\textwidth]{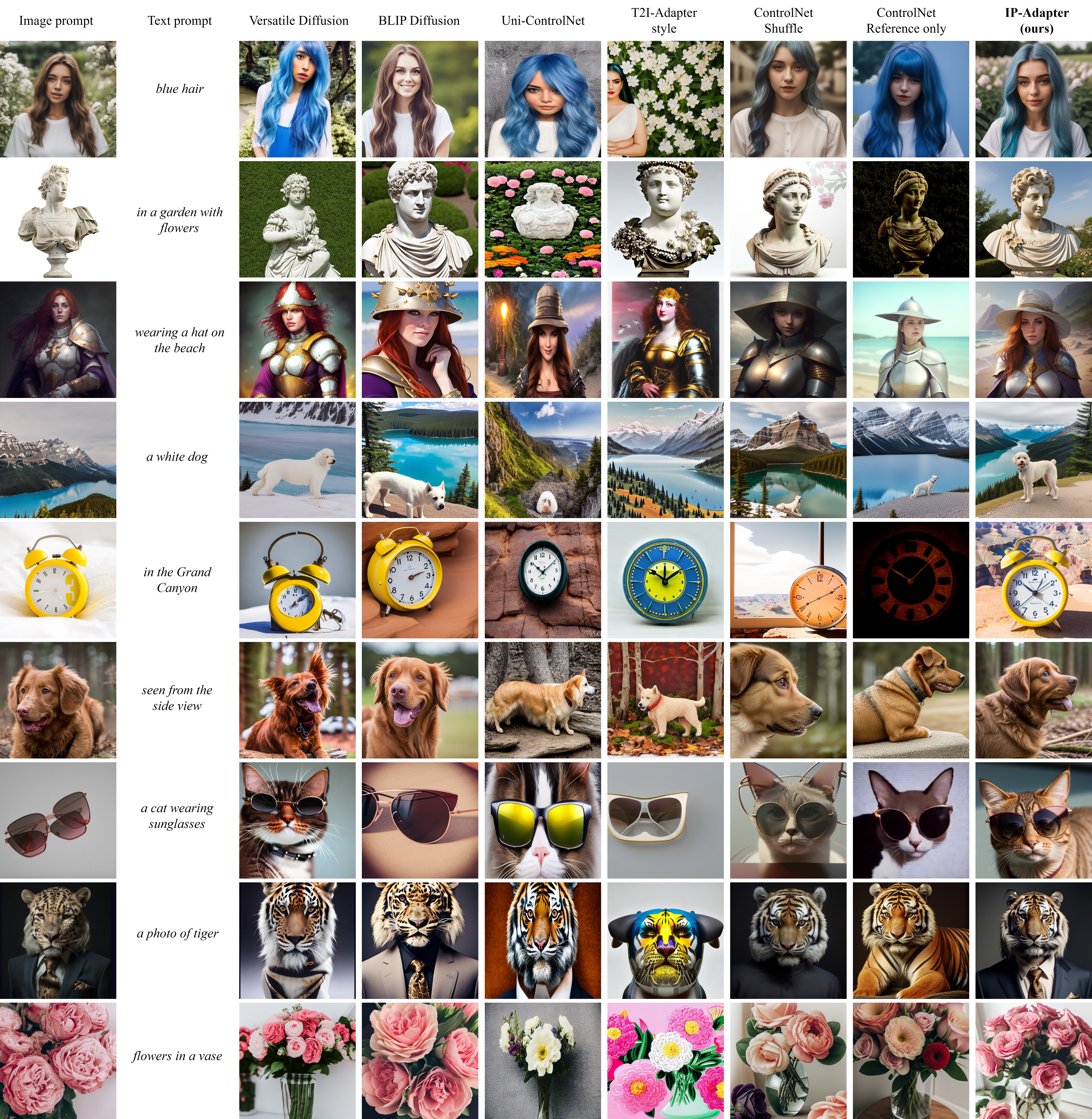}
     \vspace{-10pt}
    \caption{Comparison with multimodal prompts between our IP-Adapter with other methods.}
    \label{fig:result7}
    \vspace{-13pt}
\end{figure*}
\renewcommand{\dblfloatpagefraction}{.8}
\begin{figure*}[!h]
    \centering
    \setlength{\abovecaptionskip}{0.5cm}
    \includegraphics[width=1.0\textwidth]{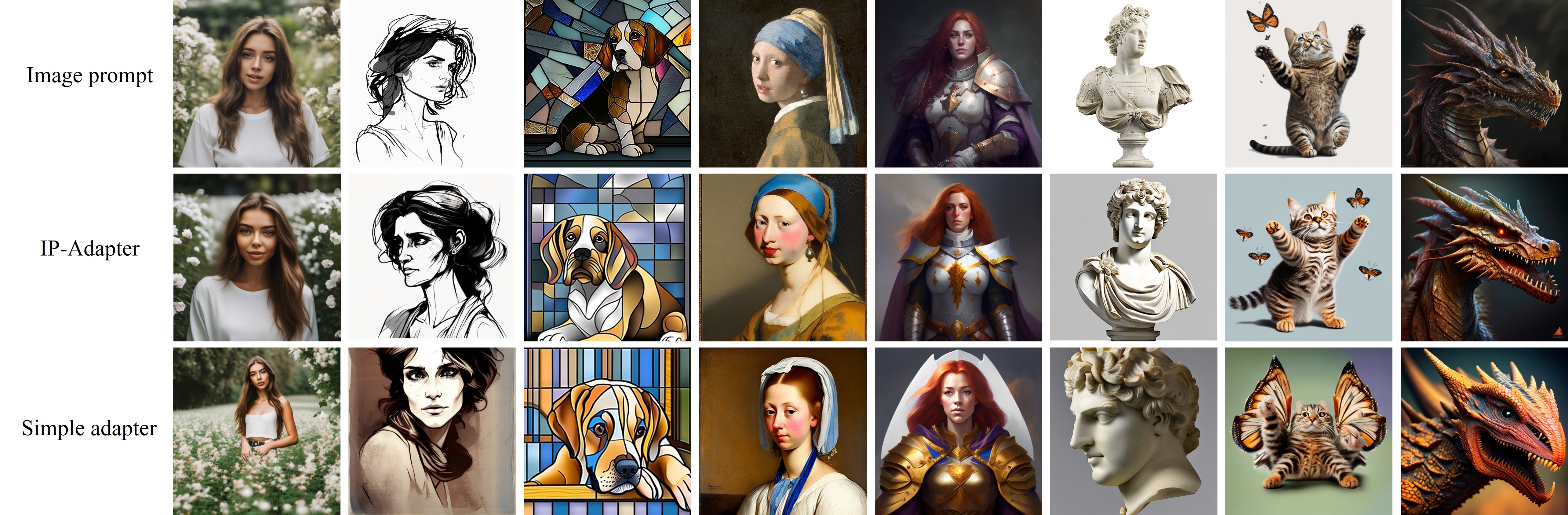}
     \vspace{-10pt}
    \caption{Comparison results of our IP-Adapter with simple adapter. The decoupled cross-attention strategy is not used in the simple adapter.}
    \label{fig:result8}
    \vspace{-13pt}
\end{figure*}
\renewcommand{\dblfloatpagefraction}{.8}
\begin{figure*}[!h]
    \centering
    \setlength{\abovecaptionskip}{0.5cm}
    \includegraphics[width=0.85\textwidth]{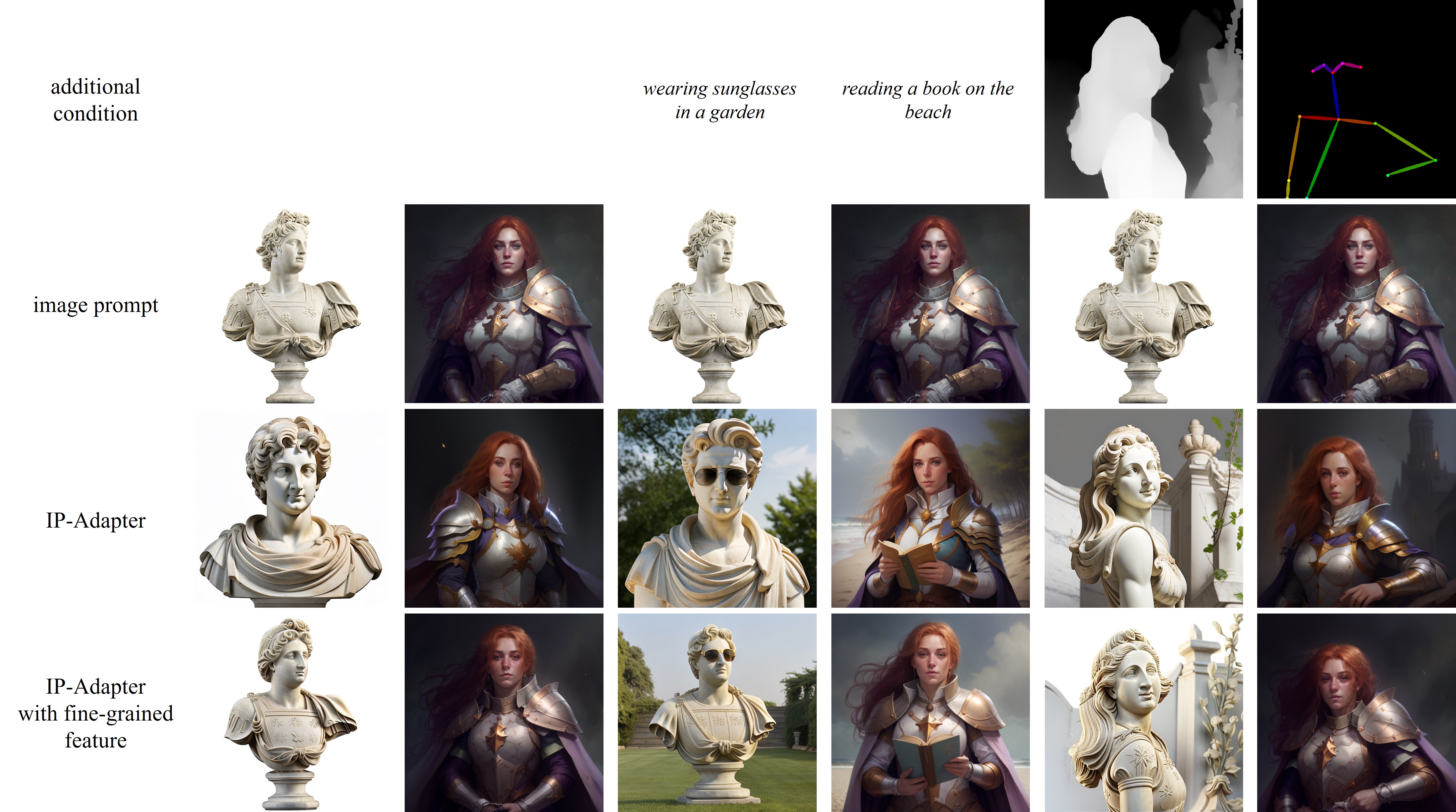}
     \vspace{-10pt}
    \caption{The visual difference of generated samples between the IP-Adapter with global features and the IP-Adapter with fine-grained features.
}
    \label{fig:result9}
    \vspace{-13pt}
\end{figure*}

\subsubsection{Structure Control}
For text-to-image diffusion models, a popular application is that we can create images with additional structure control. As our adapter does not change the original network structure, we found that the IP-Adapter is fully compatible with existing controllable tools. As a result, we can also generate controllable images with image prompt and additional conditions. Here, we combine our IP-Adapter with two existing controllable tools, ControlNet and T2I-Adapter. Figure \ref{fig:result3} shows various samples that are generated with image prompt and different structure controls: the samples of the first two rows are generated with ControlNet models, while the samples in the last row are generated with T2I-Adapters. Our adapter effectively works with these tools to produce more controllable images without fine-tuning.

We also compare our adapter with other adapters on the structural control generation, the results are shown in Figure \ref{fig:result4}. For T2I-Adapter and Uni-ControlNet, we use the default composable multi-conditions. For SeeCoder and our IP-Adapter, we use ControlNet to achieve structural control. For ControlNet Shuffle and ControlNet Reference-only, we use multi-ControlNet. As we can see, our method not only outperforms other methods in terms of image quality, but also produces images that better align with the reference image.
\subsubsection{Image-to-Image and Inpainting}
Apart from text-to-image generation, text-to-image diffusion models also can achieve text-guided image-to-image and inpainting with SDEdit~\cite{meng2021sdedit}. As demonstrated in Figure \ref{fig:result5}, we can also obtain image-guided image-to-image and inpainting by simply replacing text prompt with image prompt.
\subsubsection{Multimodal Prompts}
For the fully fine-tuned image prompt models, the original text-to-image ability is almost lost. However, with the proposed IP-Adapter, we can generate images with multimodal prompts including image prompt and text prompt. We found that this capability performs particularly well on community models. In the inference stage with multimodal prompts, we adjust $\lambda$ to make a balance between image prompt and text prompt. Figure \ref{fig:result6} displays various results with multimodal prompts using Realistic Vision V4.0 model. As we can see, we can use additional text prompt to generate more diverse images. For instance, we can edit attributes and change the scene of the subject conditioned on the image prompt using simple text descriptions.

We also compare our IP-Adapter with other methods including Versatile Diffusion, BLIP Diffusion~\cite{li2023blip}, Uni-ControlNet, T2I-Adapter, ControlNet Shuffle, and ControlNet Reference-only. The comparison results are shown in Figure \ref{fig:result7}. Compared with other existing methods, our method can generate superior results in both image quality and alignment with multimodal prompts.

\subsection{Ablation Study}
\subsubsection{Importance of Decoupled Cross-Attention}
In order to verify the effectiveness of the decoupled cross-attention strategy, we also compare a simple adapter without decoupled cross-attention: image features are concatenated with text features, and then embedded into the pretrained cross-attention layers. For a fair comparison, we trained both adapters for 200,000 steps with the same configuration. Figure \ref{fig:result8} provides comparative examples with the IP-Adapter with decoupled cross-attention and the simple adapter. As we can observe, the IP-Adapter not only can generate higher quality images than the simple adapter, but also can generate more consistent images with image prompts.
\subsubsection{Comparison of Fine-grained Features and Global Features}
Since our IP-Adapter utilizes the global image embedding from the CLIP image encoder, it may lose some information from the reference image. Therefore, we design an IP-Adapter conditioned on fine-grained features. First, we extract the grid features of the penultimate layer from the CLIP image encoder. Then, a small query network is used to learn features. Specifically, 16 learnable tokens are defined to extract information from the grid features using a lightweight transformer model. The token features from the query network serve as input to the cross-attention layers.

The results of the two adapters are shown in Figure \ref{fig:result9}. Although the IP-Adapter with finer-grained features can generate more consistent images with image prompt, it can also learn the spatial structure information, which may reduce the diversity of generated images. However, additional conditions, such as text prompt and structure map, can be combined with image prompt to generate more diverse images. For instance, we can synthesize novel images with the guidance of additional human poses.

\section{Conclusions and Future Work}
In this work, we propose IP-Adapter to achieve image prompt capability for the pretrained text-to-image diffusion models. The core design of our IP-Adapter is based on a decoupled cross-attention strategy, which incorporates separate cross-attention layers for image features. Both quantitative and qualitative experimental results demonstrate that our IP-Adapter with only 22M parameters performs comparably or even better than some fully fine-tuned image prompt models and existing adapters. Furthermore, our IP-Adapter, after being trained only once, can be directly integrated with custom models derived from the same base model and existing structural controllable tools, thereby expanding its applicability. More importantly, image prompt can be combined with text prompt to achieve multimodal image generation.

Despite the effectiveness of our IP-Adapter, it can only generate images that resemble the reference images in content and style. In other words, it cannot synthesize images that are highly consistent with the subject of a given image like some existing methods, e.g., Textual Inversion~\cite{gal2022image} and DreamBooth~\cite{ruiz2023dreambooth}. In the future, we aim to develop more powerful image prompt adapters to enhance consistency.

\bibliographystyle{unsrt}  
\bibliography{references}

\end{document}